\documentclass[a4paper, 11pt, twoside]{Thesis}  
\graphicspath{{./Figures/}}  
\setchapterpath{./Chapters/}
\setappendixpath{./Appendices/}

\usepackage[square, numbers, comma, sort&compress]{natbib}  
\usepackage{verbatim}  
\usepackage{verbatimbox}
\usepackage[usenames]{xcolor}
\usepackage{tabularx}
\usepackage{multirow}
\usepackage{wrapfig}
\usepackage{framed}
\usepackage{amssymb}
\usepackage{amsmath}
\usepackage{amsthm}
\usepackage{stmaryrd}
\usepackage{framed}
\usepackage{hyperref}
\usepackage{todonotes}
\reversemarginpar
\presetkeys{todonotes}{inline, size=\tiny, prepend, caption={TODO}}{}
\usepackage{listings}
\usepackage{longtable}
\usepackage{multicol}
\usepackage[multiple]{footmisc}
\usepackage [english]{babel}
\usepackage [autostyle, english = american]{csquotes}
\MakeOuterQuote{"}

\setboolean  {proposal}{false} 
\university  {}{http://www.uni-saarland.de/}
\department  {}{http://www.cs.uni-saarland.de/}
\group       {Universit\"at des Saarlandes}{http://www.cs.uni-saarland.de/}
\faculty     {Bosch Center for Artificial Intelligence}{https://www.bosch-ai.com/}
\supervisor  {Prof. Dr. Dietrich Klakow}
\advisor     {Dr. Heike Adel\\
              Dr. Mohamed Gad-Elrab\\
              MSc. Dragan Milchevski}
\examiner    {Prof. Dr. Dietrich Klakow}
\secondexaminer {Dr. Volha Petukhova}
\degree      {Master of Science}
\thesiskind  {Master's Thesis in Computer Science}
\authors     {Hassan Soliman}
\thesistitle {Cross-Domain Neural Entity Linking}
\date        {February 1, 2022}

\doifproposal{
	\let\thesisnameproposal\thesisname
	\thesiskind{Proposal for a \thesisnameproposal}
}

\begin{document}
\frontmatter	  

\maketitle

\doifnotproposal{
\begin{declaration}
  \begin{center}
    \bf \large
    Eidesstattliche Erkl\"{a}rung
  \end{center}
  Ich erkl\"{a}re hiermit an Eides Statt,
  dass ich die vorliegende Arbeit selbstst\"{a}ndig verfasst und keine
  anderen als die angegebenen Quellen und Hilfsmittel verwendet habe.

  \begin{center}
    \bf \large
    Statement in Lieu of an Oath
  \end{center}
  I hereby confirm that I have written this thesis on my own and 
  that I have not used any other media or materials than the ones
  referred to in this thesis.

  \vfill
  \begin{center}
    \bf \large
    Einverst\"{a}ndniserkl\"{a}rung
  \end{center}
  Ich bin damit einverstanden, dass meine (bestandene) Arbeit in beiden 
  Versionen in die Bibliothek der Informatik aufgenommen und damit 
  ver\"{o}ffentlicht wird. 

  \begin{center}
    \bf \large
    Declaration of Consent
  \end{center}
  I agree to make both versions of my thesis (with a passing grade) 
  accessible to the public by having them added to the library of the
  Computer Science Department.

  \vfill
  \vfill
   
  Datum/Date: February 1, 2022\\ 
  \rule[1em]{25em}{0.5pt}  

  Unterschrift/Signature: \\ 
  \rule[1em]{25em}{0.5pt}  
   
\end{declaration}
  \cleardoublepage  
}

\begin{abstract_content}

\textit{Entity Linking} is the task of matching a mention to an entity in a given knowledge base (KB). It contributes to annotating a massive amount of documents existing on the Web to harness new facts about their matched entities. However, existing \textit{Entity Linking} systems focus on developing models that are typically domain-dependent and robust only to a particular knowledge base on which they have been trained. The performance is not as adequate when being evaluated on documents and knowledge bases from different domains. 

Approaches based on pre-trained language models, such as Wu \textit{et al.} (2020)~\cite{wu_scalable_2020}, attempt to solve the problem using a zero-shot setup, illustrating some potential when evaluated on a general-domain KB. Nevertheless, the performance is not equivalent when evaluated on a domain-specific KB. To allow for more accurate \textit{Entity Linking} across different domains, we propose our framework: \textit{Cross-Domain Neural Entity Linking} (CDNEL). Our objective is to have a single system that enables simultaneous linking to both the general-domain KB and the domain-specific KB. CDNEL works by learning a joint representation space for these knowledge bases from different domains. It is evaluated using the external \textit{Entity Linking} dataset (Zeshel) constructed by Logeswaran \textit{et al.} (2019)~\cite{logeswaran_zero-shot_2019} and the Reddit dataset collected by Botzer \textit{et al.} (2021)~\cite{botzer_reddit_2021}, to compare our proposed method with the state-of-the-art results. The proposed framework uses different types of datasets for fine-tuning, resulting in different model variants of CDNEL. When evaluated on four domains included in the Zeshel dataset, these variants achieve an average precision gain of 9\%.

\end{abstract_content}

\doifnotproposal{

\begin{acknowledgements}
\setcode{utf8}
\begin{arabtext}
الْحَمْدُ لِلَّهِ حَمْدًا كَثِيرًا طَيِّبًا مُبَارَكًا فِيهِ
\end{arabtext}

This work would not have been possible without the support of many people. Many thanks to my advisors, Dr. Heike Adel, Dr. Mohamed Gad-Elrab, and MSc. Dragan Milchevski, who read my numerous revisions and gave valuable feedback. I am wholeheartedly grateful to them for their great advice and guidance on this Master's thesis. I would like to express my gratitude to Prof. Dr. Dietrich Klakow for giving me the opportunity to work under his supervision. I would like to sincerely thank Dr. Volha Petukhova for offering to review my Master's thesis. I am grateful to the entire Bosch Center for Artificial Intelligence (BCAI) family for their support and encouragement throughout the Master's thesis.

Apart from the Master's thesis, I would like to express my sincere appreciation to Gad for always giving me helpful advice and solutions to the problems I face both on a personal and educational level. He is always willing to share his valuable expertise at all stages of my journey since I moved to Saarbrücken. Working with him, Heike and Dragan helps me learn essential research skills that I believe will significantly help me in my future career. I hope I will have the opportunity to work with them again in the future.

I cannot find words to express how much I am indebted to my mother. She is the backbone of my life. I am grateful to her, my father, my sister Shrouk, my brother Eslam, my uncles and my aunt for their sincere support and prayers throughout my life. Their support is a major factor that has motivated me throughout my long educational journey. I genuinely appreciate that I can always rely on them and that they have so much patience with my constant absence. I would like to sincerely thank my fiancée Zena from the bottom of my heart for always supporting me, especially in difficult times. I am grateful for her patience during the last months. 

A heartfelt thanks goes to Nedal who has always motivated me even when we are abroad most of the time. I would like to dedicate special thanks to Abdallah and Magdy for their constant support, both personally and professionally. I would like to thank my second family of friends in Saarbrücken, Adeeb, Afnan, Aleen, Amr, Dalia, Fatima, Mario, Michael, Mostafa, Osama, Taha, Youssef and everyone who is close to me and was always there when I needed their support. I believe I am blessed to be surrounded by all these intelligent, caring and enthusiastic personalities.

Finally, I would like to thank everyone who pledged their support or made dua'a for me.

\end{acknowledgements}
\cleardoublepage
}


\tableofcontents
\clearpage

\doifnotproposal{
  \thispagestyle{empty}  
  \dedicatory{Dedicated to my uncle \textbf{Hassan Elshrief}, who is no longer with us. He will always be in my heart. I will never forget how much he helped me in my life. He is one of the main reasons I could do this Master's in Germany. May Allah have mercy on his soul.}
}

\mainmatter	  


\loadchapter{Introduction}{Introduction}
\loadchapter{Background}{Background}
\loadchapter{Related_Work}{Related Work}
\loadchapter{Datasets}{Datasets}
\loadchapter{Methodology}{Methodology}
\loadchapter{Experimental_Evaluation}{Experimental Evaluation}
\loadchapter{Conclusion}{Conclusion}

\clearpage

\appendix

\backmatter

\doifnotproposal{
	\lhead{\emph{List of Figures}}
	\addtotoc{List of Figures}
	\listoffigures
	\clearpage

  	\addtotoc{List of Tables}
  	\lhead{\emph{List of Tables}}
	\listoftables
	\clearpage
	
}


\label{Bibliography}
\addtotoc{Bibliography}
\lhead{\emph{Bibliography}}  
\bibliographystyle{unsrtnat}  
\bibliography{Bibliography}  

\end{document}